\documentclass{article}

\PassOptionsToPackage{numbers, compress}{natbib}
\usepackage[preprint]{neurips_2019}
\usepackage{neurips_2019}

\usepackage{amsmath}
\usepackage[utf8]{inputenc} %
\usepackage[T1]{fontenc}    %
\usepackage{hyperref}       %
\usepackage{url}            %
\usepackage{booktabs}       %
\usepackage{amsfonts}       %
\usepackage{nicefrac}       %
\usepackage{microtype}      %
\usepackage{graphicx}
\usepackage{color}

\usepackage{multirow,tabularx}
\usepackage{caption}
\usepackage{subcaption}

\title{Self-supervised Training of Proposal-based Segmentation via Background Prediction}

\author{%
  Isinsu Katircioglu$^1$, Helge Rhodin$^1$, Victor Constantin$^1$, Jörg Spörri$^2$,\\ \bf Mathieu Salzmann$^1$, and  Pascal Fua$^1$\\
  \texttt{isinsu.katircioglu@epfl.ch}\\\\
  \and
  $^1$Computer Vision Laboratory\\
  EPFL\\
  Lausanne, Switzerland
  \and
  $^2$Sports Medical Research Group\\
  Balgrist University Hospital, University of Zurich\\
  Zurich, Switzerland
}

\begin{document}

\newif\ifdraft
\drafttrue

\ifdraft

\newcommand{\VC}[1]{{\color{blue}{\bf vc: #1}}}
\newcommand{\vc}[1]{{\color{blue} #1}}

\newcommand{\JS}[1]{{\color{cyan}{\bf js: #1}}}
\newcommand{\NEW}[1]{{\color{red}{#1}}}

\else

\newcommand{\VC}[1]{{\color{green}{}}}

\newcommand{\NEW}[1]{#1}
\fi

\newcommand{\va}{\mathbf{a}}
\newcommand{\vb}{\mathbf{b}}
\newcommand{\vcc}{\mathbf{c}}
\newcommand{\vd}{\mathbf{d}}
\newcommand{\ve}{\mathbf{e}}
\newcommand{\vf}{\mathbf{f}}
\newcommand{\vg}{\mathbf{g}}
\newcommand{\vh}{\mathbf{h}}
\newcommand{\vi}{\mathbf{i}}
\newcommand{\vj}{\mathbf{j}}
\newcommand{\vk}{\mathbf{k}}
\newcommand{\vl}{\mathbf{l}}
\newcommand{\vm}{\mathbf{m}}
\newcommand{\vn}{\mathbf{n}}
\newcommand{\vo}{\mathbf{o}}
\newcommand{\vp}{\mathbf{p}}
\newcommand{\vq}{\mathbf{q}}
\newcommand{\vr}{\mathbf{r}}
\newcommand{\vt}{\mathbf{t}}
\newcommand{\vu}{\mathbf{u}}
\newcommand{\vv}{\mathbf{v}}
\newcommand{\vw}{\mathbf{w}}
\newcommand{\vx}{\mathbf{x}}
\newcommand{\vy}{\mathbf{y}}
\newcommand{\vz}{\mathbf{z}}

\newcommand{\mA}{\mathbf{A}}
\newcommand{\mB}{\mathbf{B}}
\newcommand{\mC}{\mathbf{C}}
\newcommand{\mD}{\mathbf{D}}
\newcommand{\mE}{\mathbf{E}}
\newcommand{\mF}{\mathbf{F}}
\newcommand{\mG}{\mathbf{G}}
\newcommand{\mH}{\mathbf{H}}
\newcommand{\mI}{\mathbf{I}}
\newcommand{\mJ}{\mathbf{J}}
\newcommand{\mK}{\mathbf{K}}
\newcommand{\mL}{\mathbf{L}}
\newcommand{\mM}{\mathbf{M}}
\newcommand{\mN}{\mathbf{N}}
\newcommand{\mO}{\mathbf{O}}
\newcommand{\mP}{\mathbf{P}}
\newcommand{\mQ}{\mathbf{Q}}
\newcommand{\mR}{\mathbf{R}}
\newcommand{\mS}{\mathbf{S}}
\newcommand{\mT}{\mathbf{T}}
\newcommand{\mU}{\mathbf{U}}
\newcommand{\mV}{\mathbf{V}}
\newcommand{\mW}{\mathbf{W}}
\newcommand{\mX}{\mathbf{X}}
\newcommand{\mY}{\mathbf{Y}}
\newcommand{\mZ}{\mathbf{Z}}

\newcommand{\cA}{\mathcal A}
\newcommand{\cB}{\mathcal B}
\newcommand{\cC}{\mathcal C}
\newcommand{\cD}{\mathcal D}
\newcommand{\cE}{\mathcal E}
\newcommand{\cF}{\mathcal F}
\newcommand{\cG}{\mathcal G}
\newcommand{\cH}{\mathcal H}
\newcommand{\cI}{\mathcal I}
\newcommand{\cJ}{\mathcal J}
\newcommand{\cK}{\mathcal K}
\newcommand{\cL}{\mathcal L}
\newcommand{\cM}{\mathcal M}
\newcommand{\cN}{\mathcal N}
\newcommand{\cO}{\mathcal O}
\newcommand{\cP}{\mathcal P}
\newcommand{\cQ}{\mathcal Q}
\newcommand{\cR}{\mathcal R}
\newcommand{\cS}{\mathcal S}
\newcommand{\cT}{\mathcal T}
\newcommand{\cU}{\mathcal U}
\newcommand{\cV}{\mathcal V}
\newcommand{\cW}{\mathcal W}
\newcommand{\cX}{\mathcal X}
\newcommand{\cY}{\mathcal Y}
\newcommand{\cZ}{\mathcal Z}

\newcommand{\TODO}[1]{\textcolor{cyan}{#1}}

\newcommand{\ST}{\mathcal{T}}
\newcommand{\SST}{\mathcal{T}_S}

\newcommand{\R}{\mathbb{R}}
\newcommand{\Seg}{\mathbf{S}} %
\newcommand{\Latent}{\mathbf{L}}
\newcommand{\LatentG}{\Latent^{\text{3D}}} %
\newcommand{\LatentA}{\Latent^\text{app}} %
\newcommand{\LatentBG}{\mB} %

\newcommand{\norm}[1]{\left\lVert#1\right\rVert}
\newcommand{\argmin}{\operatornamewithlimits{argmin}}
\newcommand{\erf}{\operatornamewithlimits{erf}}

\newcommand{\Var}{\operatornamewithlimits{Var}}

\newcommand{\parag}[1]{\vspace{-3mm}\paragraph{#1}}

\newcommand{\handheld}[0]{{\bf Handheld190k}}
\newcommand{\ski}[0]{{\bf Ski-PTZ-Dataset}}
\newcommand{\human}[0]{{\bf H36M}}

\newcommand{\ours}[0]{{\bf Ours}}
\newcommand{\direct}[0]{{\bf Resnet}}
\newcommand{\LCR}[0]{{\bf LCR}}
\newcommand{\ECCV}[0]{{\bf NVS-encoder}}
\newcommand{\CVPR}[0]{{\bf Multiview}}
\newcommand{\auto}[0]{{\bf Auto-encoder}}

\maketitle

\begin{abstract}

While supervised object detection methods achieve impressive accuracy, they generalize poorly to images whose appearance significantly differs from the data they have been trained on. To address this in scenarios where annotating data is prohibitively expensive, we introduce a self-supervised approach to object detection and segmentation, able to work with monocular images captured with a moving camera. At the heart of our approach lies the observation that segmentation and background reconstruction are linked tasks, and the idea that, because we observe a structured scene, background regions can be re-synthesized from their surroundings, whereas regions depicting the object cannot.
We therefore encode this intuition as a self-supervised loss function that we exploit to train a proposal-based segmentation network. To account for the discrete nature of object proposals, we develop a Monte Carlo-based training strategy that allows us to explore the large space of object proposals. Our experiments demonstrate that our approach yields accurate detections and segmentations in images that visually depart from those of standard benchmarks, outperforming existing self-supervised methods and  approaching weakly supervised ones that exploit large annotated datasets.

\end{abstract}

\section{Introduction}

Recent object detection and segmentation methods have reached impressive precision and recall rates when trained and tested on large annotated datasets \cite{Lin14a}. However, large and varied datasets do not warrant the best possible performance in a particular application domain, as each comes with its own challenges and opportunities. While domain-specific models are thus needed, it is impractical to annotate separate and sufficiently large datasets for deep learning.

Therefore, weakly- and self-supervised detection and segmentation of salient foreground objects in complex scenes has recently gained attention~\cite{Croitoru18,Eslami16,Crawford19,Rhodin19}. These methods promise effortless processing of community videos with little human intervention. However, a closer look at existing techniques reveals that they make strong assumptions ranging from target objects being on top of static background or relying on pre-trained object localization, object-boundary detection, and optical flow networks. This severely limits their applicability in practice. 

To develop a more generic technique, we start from the observation that in most images the background forms a consistent, natural scene. Therefore the appearance of any background patch can be predicted from its surroundings. By contrast, a salient object's appearance is unpredictable from the neighboring scene content and can be expected to be very different from what an inpainting algorithm would produce. We incorporate this insight into a proposal-generating deep network whose architecture is inspired by those of YOLO~\cite{Redmon16} and MaskRCNN~\cite{He17} but does not require explicit supervision.

For each proposal, we synthesize a background image by masking out the corresponding region and inpainting it from the remaining image. The loss function we minimize favors the largest possible distance between this reconstructed background and the input image. This encourages the network to select regions that cannot be explained from their surrounding and are therefore salient. To handle the discrete nature of the proposals, we introduce a Monte Carlo-based strategy to train our network. It operates on a discrete distribution, is unbiased, exhibits low variance, and is end-to-end trainable. 

\begin{figure}
  \centering
\resizebox{1\linewidth}{!}{%
 \setlength{\tabcolsep}{1px}
  \begin{tabular}{cccc}%
  \includegraphics[width=0.25\textwidth]{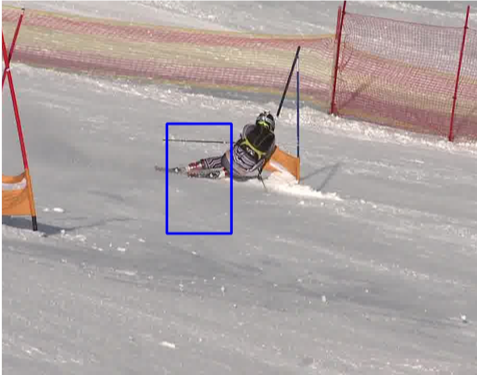} & 
  \includegraphics[width=0.25\textwidth]{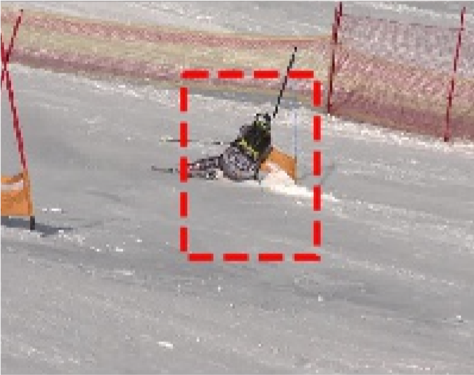} & 
  \includegraphics[width=0.25\textwidth]{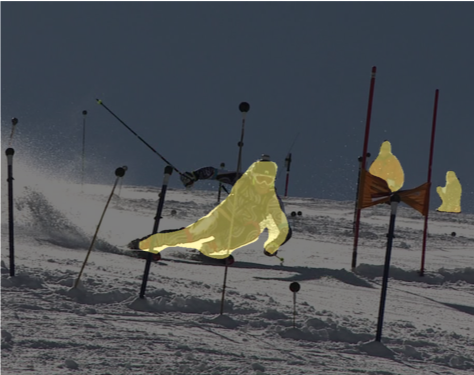} & 
  \includegraphics[width=0.25\textwidth]{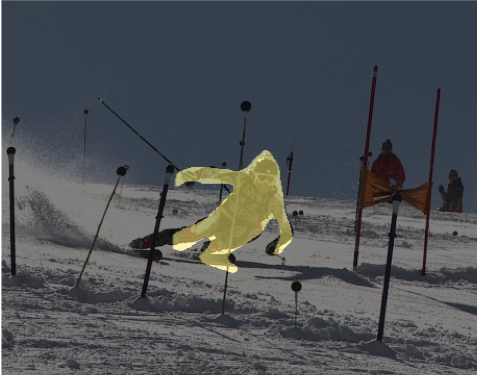} \\ 
  {\small (a) YOLOv3} & {\small (b) Ours }& {\small (c) Mask-RCNN} & {\small (d) Ours}
  \end{tabular}
}
  \caption{\textbf{Domain specific detection examples.} 
  Our self-supervised method detects the skier well, while YOLO trained on a general dataset does not generalize to this challenging domain. We also compare to MaskRCNN, which succeeds on the skier but detects false positives.
  }
  \label{fig:teaser}
\end{figure}

We demonstrate the effectiveness of our unsupervised method on several datasets captured with increasingly mobile cameras, ranging from static to pan-tilt-zoom and hand-held. We will show that our approach applies to images acquired in conditions significantly more general than those of standard benchmarks, without requiring \emph{any} manual annotation. Thus, as shown in Fig.~\ref{fig:teaser}, it approaches the quality and sometimes outperforms state-of-the-art detectors that have been trained on large annotated datasets. Retraining or fine-tuning these methods on this data could be done but would require supervision that is hard to obtain, which makes a self-supervised approach attractive. We will make our code and \handheld{} dataset publicly available upon acceptance of the paper.

\section{Related Work}
\label{sec:related}

Most salient object detection and segmentation algorithms are fully-supervised \cite{Redmon16,He17,Song18,Cheng17} and require large annotated datasets with paired images and labels. Our goal is a purely self-supervised method that succeed without segmentation and object bounding box annotations. This is different from those methods requiring domain-specific annotation at training but not at test time, which are often also referred to as  \emph{unsupervised object detection} methods \cite{Hu18}. We focus our discussion on neural methods and refer to \cite{Koh17} for the discussion of methods using hand-crafted optimization.

\parag{Weakly-supervised methods.}
A classical weakly-supervised example is the Hough Matching algorithm ~\citep{Cho15}. It uses an object classification dataset and identifies foreground as the image regions that have re-occurring Hough features within images of the same class.
Similar principles have been followed using deep networks trained for object detection~\citep{Wei17,Jain17}, optical flow~\cite{Tokmakov17}, and object saliency \cite{Li18x}.
These methods make the implicit assumption that the background varies across examples and can therefore be excluded as noise. This assumption is violated in the targeted case of training on domain-specific images, where foreground and background are similar across examples.

\parag{Motion-based methods}
Given video sequences, the temporal information can be exploited by assuming that the background changes slowly~\cite{Barnich11} and linearly~\cite{Stretcu15}. However, even a static scene induces non-homogeneous deformations under camera translation, and it can be difficult to handle all types of camera motion (pan, tilt, zoom) at different pastes, cuts within a single video, and distinguish articulated human motion from background motion~\cite{Russell14}. 
By iteratively refining the crude background subtraction results from~\cite{Stretcu15} with an assemble of student and teacher networks~\cite{Croitoru18,Croitoru19}, some of the resulting errors could be corrected. However, a strong dependency on the teacher used for bootstrapping remains. 

Our approach is conceptually closely related to VideoPCA, which models the background as the part of the scene that can be explained by a low-dimensional linear basis~\cite{Stretcu15}. This implicitly assumes that the foreground is harder to model than the background and can therefore be separated as the non-linear residual. Instead of motion cues, we introduce a new assumption on the predictability of image patches from the spatial neighborhood.

\parag{Self-supervised Methods.}

Most similar to our approach are the self-supervised ones to object-detection~\cite{Eslami16,Crawford19,Rhodin19} that complement auto-encoder networks by an attention mechanism. These methods first detect one or several bounding boxes whose content is extracted using a spatial transformer \cite{Jaderberg15}. This content is then passed through an auto-encoder and re-composited with a background. In~\cite{Rhodin19} the background is assumed to be static and in~\cite{Eslami16,Crawford19} even single colored, a severe restriction in practice.  Crawford et al.~\cite{Crawford19} use a proposal-based network similar to ours, but resort to approximating the proposal distribution with a continuous one to be differentiable. We demonstrate that much simpler importance sampling is sufficient. By contrast to all of these methods, our approach works with images acquired using a moving camera and given an arbitrarily colored background.

In addition to object detection, the algorithm of~\cite{Rhodin19} also returns instance segmentation masks by reasoning about the extent and depth ordering of multiple people in a multi-camera scene. However, this requires multiple static cameras and a static background at training time, as does the approach of~\cite{Baque17b} that performs instance segmentation in crowded scenes.

\section{Method}

\begin{figure*}[t]
	\begin{center}
		\includegraphics[width=0.8\linewidth]{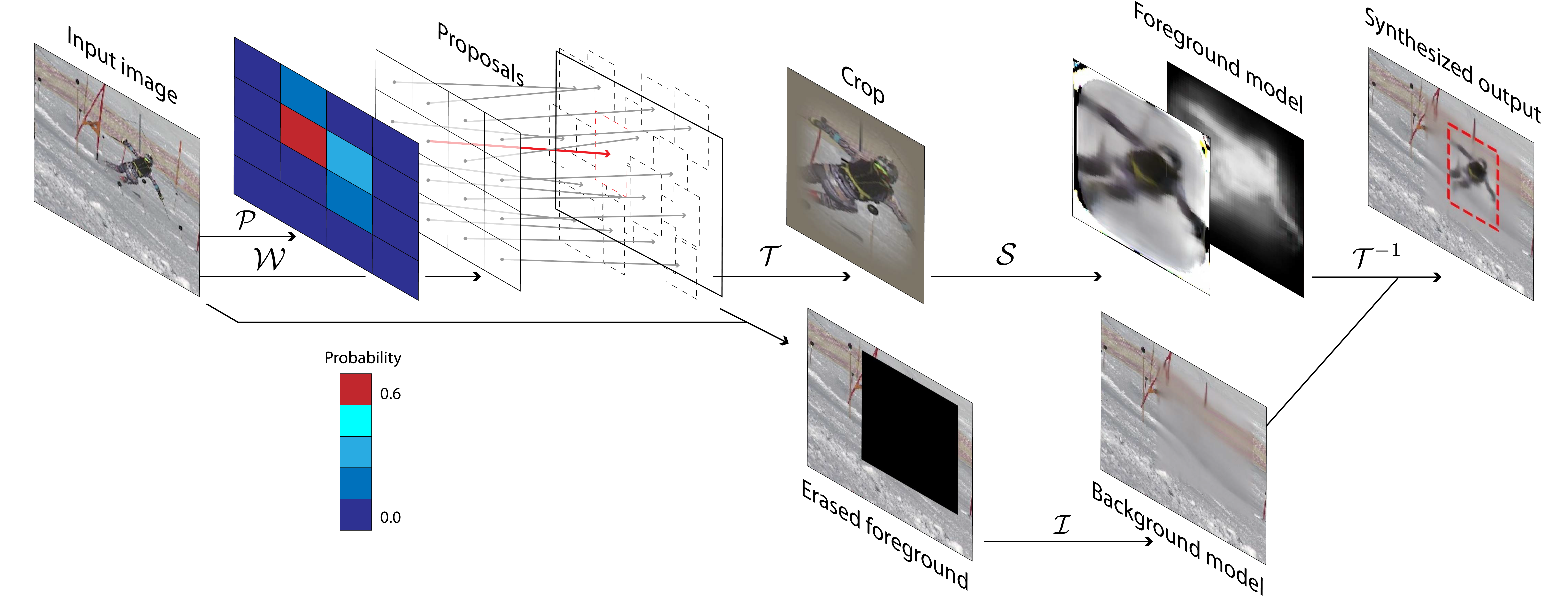}
	\end{center}
	\caption{\textbf{Method overview.} Encoder-decoder network ($\cS$) with an attention mechanism defined by proposal-based detection ($\cP,\cW$) and spatial transformers ($\cT,\cT^{-1})$. Carefully designed objective functions make it possible to train this network entirely self-supervised on unknown scenes with a moving background and hand-held camera via an inpainting network ($\cI$).}
	\label{fig:pipeline}
\end{figure*}

Our goal is to learn a salient object detector and segmentor from unlabeled videos acquired in as generic a setup as possible, including using hand-held cameras. At inference time, our algorithm takes a single image $\mI \in \R^{W \times H}$ as input and outputs a bounding box $\vb_m \in \R^{2\times2}$, in terms of center location, width, and height, and segmentation mask $\mS \in \R^{128 \times 128}$ within that window. Internally, it is a multi-stage process, as visualized in Fig.~\ref{fig:pipeline}. A first network predicts a set of $C$ candidate object locations $(\vb_{c})^C_{c=1}$ and corresponding probabilities $(p_{c})^C_{c=1}$. We use a fully-convolutional architecture similar to YOLO~\cite{Redmon16}, as used by the supervised state-of-the-art methods.
 Second, the $\vb_{m}$ with highest probability $p_m \geq p_c \; , \; \forall c$ is chosen and its content is decoded into foreground $\hat{\mI} \in \R^{128 \times 128}$, segmentation mask $\mS$, and background $\mB \in \R^{W \times H}$ with separate auto-encoder branches.

\subsection{Self-Supervised Training}

Given a set of unlabeled images $(\mI_i)^N_{i=1}$,
our goal is to train two neural networks $\cW(\mI)$ and $\cP(\mI)$ to propose suitable bounding box candidates and, respectively, the probabilities to select the $\vb=\cW(\mI)[c]$ that contains an object. Because object locations are unknown,
we do this with an autoencoder objective, $\ell\left(\cF(\mI, \vb, \mB), \mI\right)$, that measures
 how well the autoencoder $\cF$ reproduces the input image $\mI$ on top of a background $\mB$, with the attention on $\vb$. As in \cite{Crawford19,Rhodin19}, it is implemented with spatial transformer $\cT$ that crops the area of interest, followed by a bottle-neck autoencoder $\cS$ that produces foreground and segmentation mask, and second spatial transformer $\cT^{-1}$ that undoes the cropping and blends synthesized foreground and background. Formally, we write
 \begin{equation}
 \cF(\mI,\vb, \mB) = \cT^{-1}(\hat{\mI},\mS,\vb,\mB) , \text{ with } \cS(\cT(\mI,\vb)) \mapsto (\hat{\mI},\mS).
 \end{equation}

In the simplest case, $\ell$ is a least-square loss between all pixel values and the background $\mB$ is known.
Because the attention window $\vb$ selects part of the image for decoding, this loss encourages  $(\cW,\cP)$ to focus the attention on the object so as to be able to model the foreground on top of $\mB$.  Moreover, the autoencoder only approximates the actual image, which forces the segmentation mask to just contain those parts not captured by the background.

For now, we assume $\mB$ to be known and omit it in our derivations to ease readability, but we will remove this assumption later.
In this context, we derive a probabilistic formulation in which not a single, but multiple candidates can give rise to plausible reconstructions. We then reason about the expected loss across all likely candidates,
\begin{equation}
\cO(\mI) = \mE_c \left[\ell\left(\cF(\mI, \cW(\mI)[c]), \mI\right) \right], \text{ with} \ c \sim \cP(\mI),
\label{eq:expected objective}
\end{equation}
where $\mE_c$ denotes the expectation over $c$ drawn from the proposal distribution that is output by the network $\cP(\mI)$. As natural for deep learning, we optimize Eq.~\ref{eq:expected objective} with stochastic gradient descent and mini-batches on $(\mI_i)^N_{i=1}$.
Note that minimizing this objective will jointly optimize the detection networks $(\cW,\cP)$ that generate and stochastically select proposals, and the synthesis network $\cS$ that models the object appearance and segmentation mask.

Because we have a finite set of candidates, Eq.~\ref{eq:expected objective} can be expressed deterministically as a weighted sum over all candidates as
\begin{equation}
\cO(\mI) = \sum_{c=1}^{C} \cP(c|\mI) \ell\left(\cF(\mI, c), \mI\right) \;,
\end{equation}
using the shorthand $\cF(\mI, c) = \cF(\mI, \cW(\mI)[c])$.
The objective is an explicit function of $\cP$ and could be optimized by gradient descent on $\cO$. However, this sum is inefficient to evaluate in practice when $C$ is large (e.g., in our experiments $C=64$, which does not fit in memory). This was also observed by Crawford et al.~\cite{Crawford19}, who resorted to using a continuous approximation of the discrete distribution $\cP$ to facilitate end-to-end training. Here, we propose a simpler alternative, exploiting Monte Carlo and importance sampling, which provides an unbiased estimator with low variance.

\paragraph{Monte Carlo sum.} In principle, the expectation Eq.~\ref{eq:expected objective} could be estimated by sampling a small set of $J$ candidate cells from $\cP(\mI)$, for instance, one per mini-batch element, such that
\begin{equation}
\cO(\mI) \approx \frac{1}{J} \sum^J_{j=1} \ell\left(\cF(\mI, c_j), \mI\right), \text{ with } c_j \sim \cP(\mI).
\label{eq:simple sampling objective}
\end{equation}
Unfortunately, sampling from such a discrete distribution is not differentiable with respect to its parameters, which precludes end-to-end gradient-based optimization of Eq.~\ref{eq:simple sampling objective}.

\paragraph{Importance sampling.} Instead of sampling according to $\cP$, we can rewrite Eq.~\ref{eq:expected objective} to be over an arbitrary distribution $q$, by reweighting with the quotient of both distributions. That is,
\begin{align}
\cO(\mI)
&= \mE_c\left[ \frac{\cP(c|I)}{q(c)} \ell\left(\cF(\mI, c), \mI\right) \right], \text{ with } c \sim q \;.
\label{eq:expected objective importance}
\end{align}
This change of distribution and relative weighting holds for any two probability distributions, as explained in the supplementary material.
In practice, we use a mini-batch optimization, with a single sample drawn from $q$ per image, i.e., $J=1$, for estimating
\begin{align}
\cO(\mI)
\approx \frac{1}{J} \sum^{J}_{j=1} \left( \frac{\cP(c_j|I)}{q(c_j)} \ell\left(\cF(\mI, c_j), \mI\right) \right), \text{ with } c_j \sim q.
\label{eq:sampling expected objective importance}
\end{align}

While moving the distribution into the expectation sum provides differentiability, it comes with the drawback of a potentially large variance, i.e., high approximation error for small $J$. For instance, by choosing a uniform sampling distribution $\cU$, most of the uniformly drawn samples will have a low probability in $\cP$ and therefore negligible influence.
To reduce this variance, we leverage importance sampling and set the sampling distribution $q$ to be similar or equivalent to $\cP$. Note that the fraction $\frac{\cP(c|I)}{q(c)}$ cancels numerically for $q(c) = \cP(c|I)$.
However, while values cancel, their derivatives do not; the differentiability of $\cP$ is maintained.
The sampling distribution $q$ must be treated as a constant.

In practice, we select $q$ based on the current estimate of $\cP$ to reduce the variance.
To prevent division by very small values that could lead to numerical instability, we define the new distribution $q$ as
\begin{equation}
q(c) = {\cP(c | \mI)(1-C\epsilon) + \epsilon}\;.
\end{equation}
As a side effect, $\epsilon$ increases the probability that an unlikely case is chosen, which induces a form of exploration that is helpful in the early training stages of the network.
We analyze the attained variance reduction in the supplementary material.

Notably, the gradient of Eq.~\ref{eq:sampling expected objective importance} equals that of the likelihood ratio method \cite{Glynn90} used in the REINFORCE algorithm \cite{Williams92}.
In reinforcement learning terms \cite{Sutton98}, setting $q=\cP$ would correspond to a single step of on-policy learning. We refrained from motivating our derivation in this manner because the simple importance sampling rule is sufficient to explain our approach.

\subsection{Training with Dynamic Cameras}
\label{sec:separation}

Having derived an efficient training scheme for proposal-based segmentation when $\mB$ is given, we would like to reduce the more difficult moving camera scenario to the former.
This requires predicting the background image, which, in the absence of prior shape and appearance information, requires to identify and inpaint the pixels that are different from the background.

While this task entails object detection, our primary goal, the related, yet simpler inpainting task can easily be trained by removing a region and predicting it from its immediate surrounding~\cite{Pathak16,Yu2018}. A network, $\cI$, trained on this self-supervised inpainting task, would nonetheless not reconstruct foreground objects if fully removed in the input because the surrounding background gives no cues of their presence.
Detecting foreground objects can therefore be cast as finding the area $\vb$ that, when inpainted, yields the largest image reconstruction error.

To accomplish this search efficiently, we define the background objective probabilistically, similar to the foreground we write,
\begin{equation}
\cG(\mI) = -\mE_c \left[\ell(\cI(\mI, \vb),  \mI) \right], \text{ with } c \sim \cP(\mI),
\label{eq:background objective}
\end{equation}
where $\cI$ takes the image $\mI$ and the region $\vb$ to inpaint as input and $\ell$,$\cP$, and $\vb=\cW[c]$ are as before.
Note that opposed to the foreground objective, we use the negative expectation. This negative reconstruction error encourages the selection of those regions were the true image is dissimilar to the reconstructed background when minimizing Eq.~\ref{eq:background objective}.

However, as illustrated in Figure~\ref{fig:h36m ablation}(d), straight regression with a neural network or brute force search would yield trivial solutions. For instance, erasing extensively large regions, containing an object or not, will lead to higher inpainting errors, just because of the increased number of reconstructed pixels. To prevent this, we reformulate Eq~\ref{eq:background objective} so as to compute differences only within the cropped region and normalize the result by the crop area. However, this, in turn, favors locations with high error density, irrespectively of their size, as shown in Figure~\ref{fig:h36m ablation}(b).

To overcome degenerate cases, we combine the new background objective $\cG$ with the
foreground objective $\cO$ of Eq.~\ref{eq:expected objective}, substituting the known $\mB$ with learned inpainting $\cI(\mI,\vb)$. The reason behind this is that these two terms are complementary:
While $\cG$ prefers locations that cover the object neither precisely nor entirely, $\cO$ favors a tight fit over partial coverage but has a trivial solution when $\vb$ is on a background region, i.e., not covering the object and having nothing to encode.

We exploit the complementary behavior of these two terms by partitioning their influence on the individual network components. For $\cG$, we limit the gradient flow to only update $\cP$, freezing the remaining network modules $\cW$ and $\cS$. By contrast, we use $\cO$ to update only $\cW$ and $\cS$. Thereby, Eq.~\ref{eq:background objective} controls the coarse localization through detection, while Eq.~\ref{eq:expected objective} provides fine-grained regression of $\vb$. Note that this separation into coarse and fine localization is only possible with the chosen proposal-based detection framework;
direct regression to bounding boxes, as in \cite{Rhodin19}, would preclude this important separation.

\paragraph{Inpainting network.}

In principle, any off-the-shelf inpainting network trained on large and generic background datasets could be used. For instance, \cite{Pathak16,Yu2018} can produce very plausible results. However, they hallucinate objects and are therefore ill-suited to our goal.
Instead, we train $\cW$ from scratch, by reconstructing rectangular, randomly removed image regions, as shown in Figure~\ref{fig:pipeline}. This network will attempt to memorize all the images in the training set. Nevertheless, as for generic inpainting, moving objects that are independent of the surroundings's cannot be reconstructed. Note that overfitting of this network to the training set is acceptable, if not intended, as it is not needed at inference time.

\begin{figure}
  \centering
  \begin{tabular}{@{}cccc@{}}
\includegraphics[width=0.2\linewidth,height=0.13\linewidth]{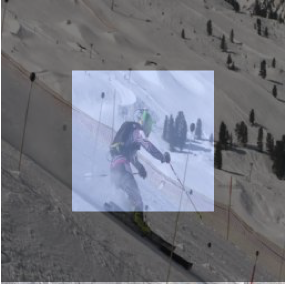}&%
\includegraphics[width=0.2\linewidth,height=0.13\linewidth]{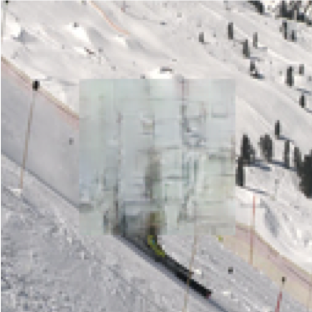}&%
\includegraphics[width=0.2\linewidth,height=0.13\linewidth]{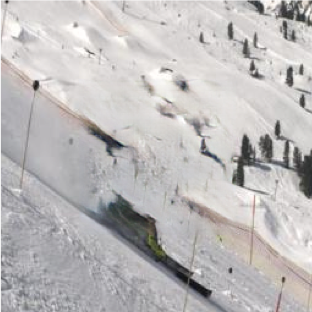}&%
\includegraphics[width=0.2\linewidth,height=0.13\linewidth]{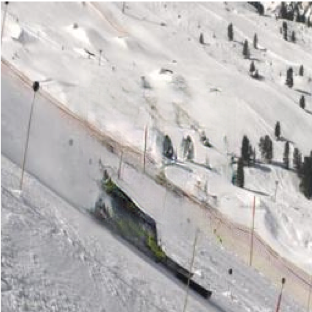}\\{\small (a)} & {\small (b) }& {\small (c) } & {\small (d) }
  \end{tabular}
  \caption{\textbf{Off-the-shelf inpainting results,} on skiing. (a) Input image with the hidden middle part, followed by inpainting with (b) \cite{Pathak16}, (c) \cite{Yu2018} trained on ImageNet. (d) and  \cite{Yu2018} trained on Places2. }
  \label{fig:inpainting others}
\end{figure}

\paragraph{Implementation details.}

Since no labels for intermediate supervision are available,
we found naive end-to-end training to be unreliable. To counteract this, we use ImageNet-trained weights for initialization;
rely on a perceptual loss on top of the per pixel $\ell_2$ loss for $\ell$; exploit the Focal Spatial Transformer (FST) of~\cite{Rhodin19} to speed up convergence; and scale the erased region in $\cI$ to be 1.1 of that predicted by $\cW$ to increase the chances of covering the object.
Moreover, we limit the location offsets to 1.5 the cell width and discard those outside the image. In addition, we rely on $\ell_2$ priors on the output of $\cP$ and $\cW$, and an $\ell_1$ prior on $\mS$. The pixel reconstruction and perceptual losses are weighted 1:2, and the priors have a weight of 0.1, 1, and 0.1, respectively, to compensate for their different magnitudes. Additional details are given in the supplemental document.

\section{Experiments}
\label{sec:eval}

In this section, we evaluate our approach to self-supervised salient object detection and segmentation. Note that our algorithm works on single images at inference time and only requires the background inpainting model at training time. Even though our approach is not people specific, we focus here on people-detection because this is the only domain for which there are benchmark datasets that contain relatively long sequences of the same scene, which is what our method requires for training purposes. 
First, we use a controlled environment with arbitrary but static background to compare our method to a state-of-the-art self-supervised one and to perform an ablation study. 
Then, we use skiing footage acquired using PTZ-cameras along with footage of people performing 14 everyday activities recorded using hand-held cameras to demonstrate that existing supervised methods that do well in the controlled environment struggle to adapt to such challenging conditions, whereas our approach delivers promising results. 
We provide additional results in the supplementary material.

\subsection{People in a Controlled Environment}

We compare our method against state-of-the-art ones on the \textbf{Human3.6m} dataset~\cite{Ionescu14a} that comprises 3.6 million frames and a set of 15 motion classes. It features nine subjects, five for training and two for validation, seen from different viewpoints against a static background and with good illumination.

\parag{Comparative Results.} On the left side of Table~\ref{tbl:results_h36m}, we compare our detection accuracy to that of a very recent self-supervised deep learning method~\cite{Rhodin19}, using the mean detection precision (mAP), the mean precision of having an intersection-over-union (IoU) of more than 50\%. Our slightly lower accuracy stems from not explicitly assuming a static background, which \cite{Rhodin19} does. It is valid in a lab but results in total failure in outdoor scenes with moving backgrounds, such as those discussed below.

\begin{figure}
  \centering
  \begin{tabular}{@{}cccc@{}}

\includegraphics[width=0.2\linewidth]{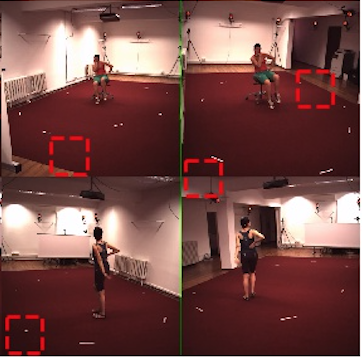} &
\includegraphics[width=0.2\linewidth]{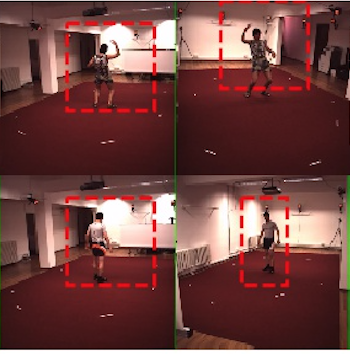} &

  \includegraphics[width=0.2\linewidth]{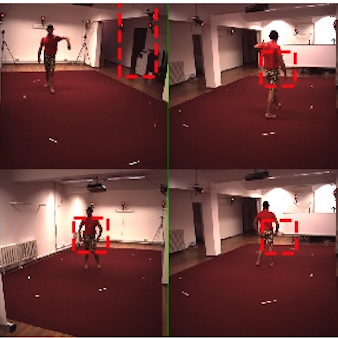} &
  \includegraphics[width=0.2\linewidth]{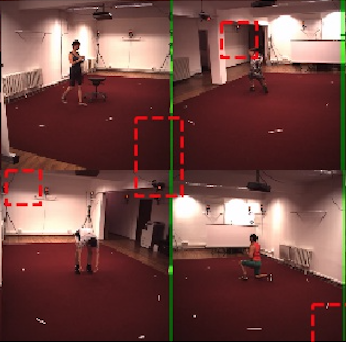}\\
  {\small (a) } & {\small (b) }& {\small (c) } & {\small (d) }\vspace{2mm} \\ \\
  \end{tabular}
  \vspace{-5mm}
  \caption{\textbf{Ablation study on H36M.} (a) Uniform sampling does not converge. (b) Joint training of $\cO$ and $\cG$ (c) only $\cG$ (d) direct regression of a single bounding box using $\cO$ and $\cG$.}
  \label{fig:h36m ablation}
\end{figure} %

\parag{Ablation Study.}
Here we show that our model choices for training and probabilistic inference are important.
Using uniform sampling instead of importance sampling, as described in Eq.~\ref{eq:sampling expected objective importance}, does not converge, as shown in Fig.~\ref{fig:h36m ablation}(a). Fig.~\ref{fig:h36m ablation}(b) shows that
joint instead of our separated training of $\cP$ and $\cW$ with $\cO$ and $\cG$ produces bounding boxes that are too large. Fig.~\ref{fig:h36m ablation}(c) shows that using only the background objective leads to small detections that miss the subject and (d) that direct regression without multiple candidates diverges. These failure cases are representative for the whole dataset.

\begin{figure}
  \centering
\resizebox{1\linewidth}{!}{%
\begin{tabular}{@{}cccccc@{}}%
\includegraphics[width=0.15\linewidth]{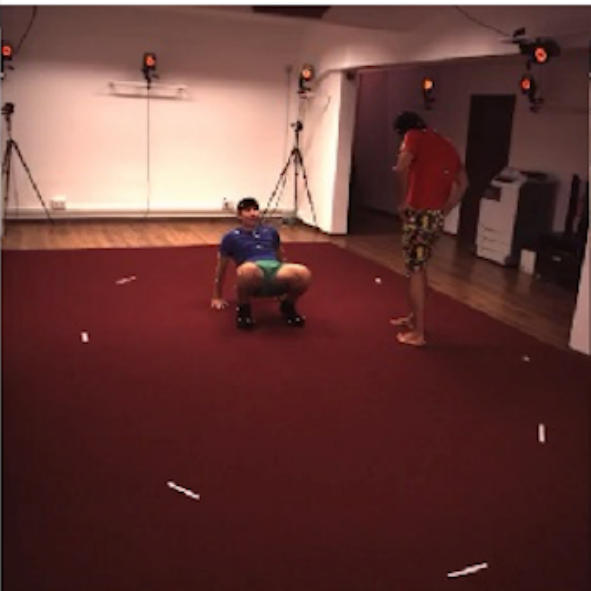} &
\includegraphics[width=0.15\linewidth]{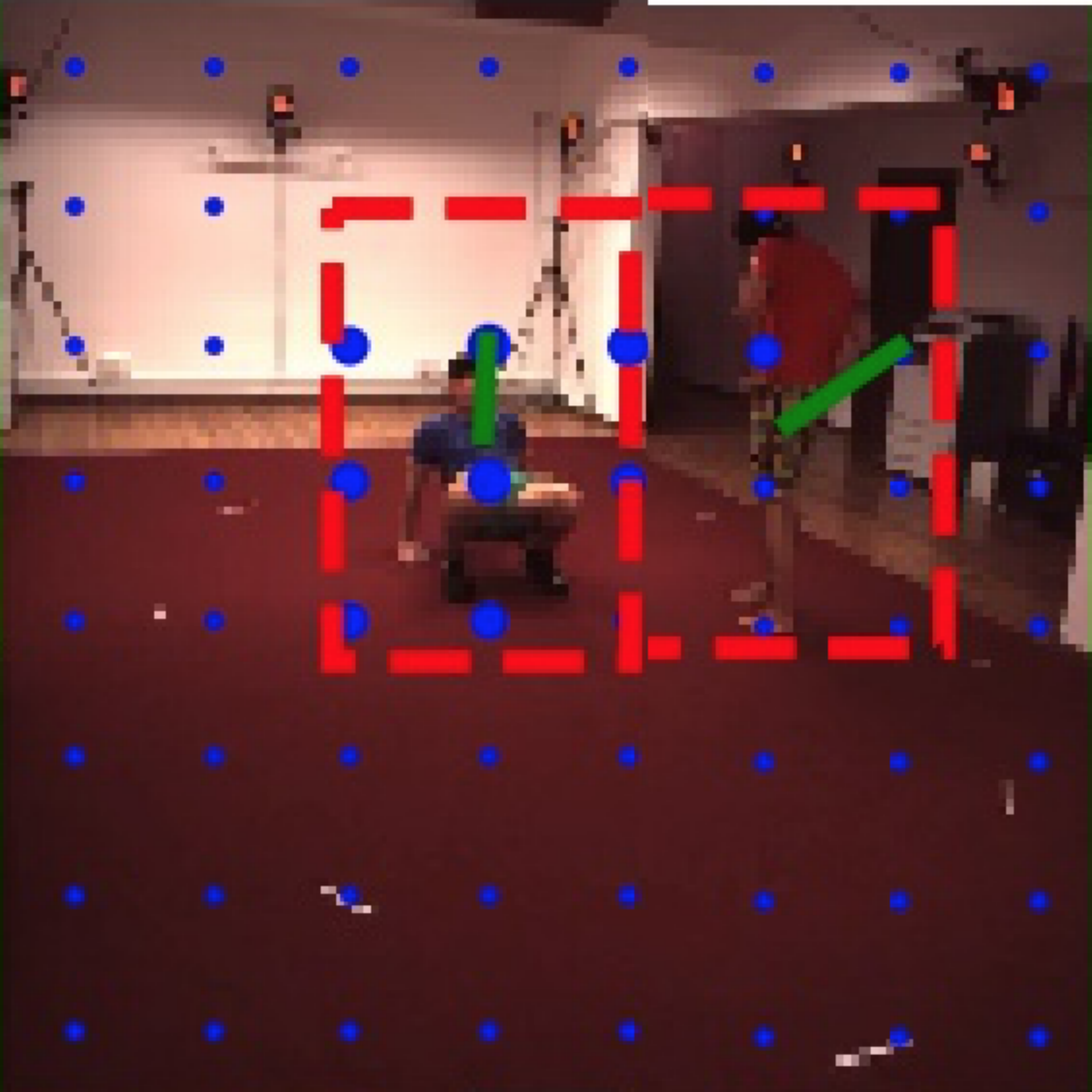} &
\includegraphics[width=0.15\linewidth]{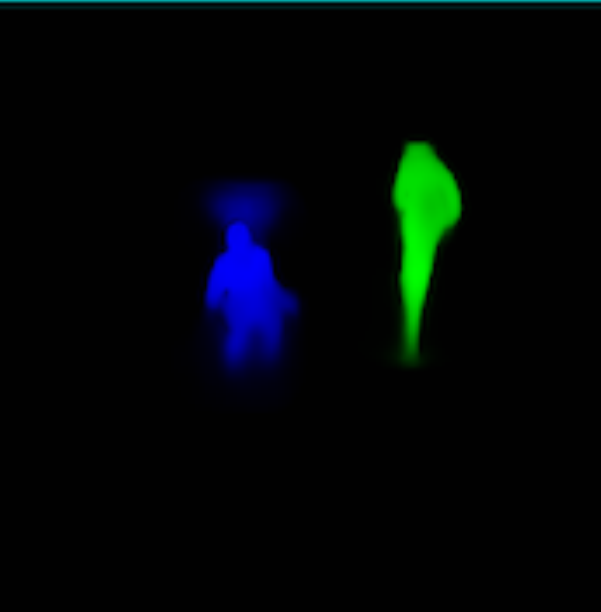} &
\includegraphics[width=0.15\linewidth]{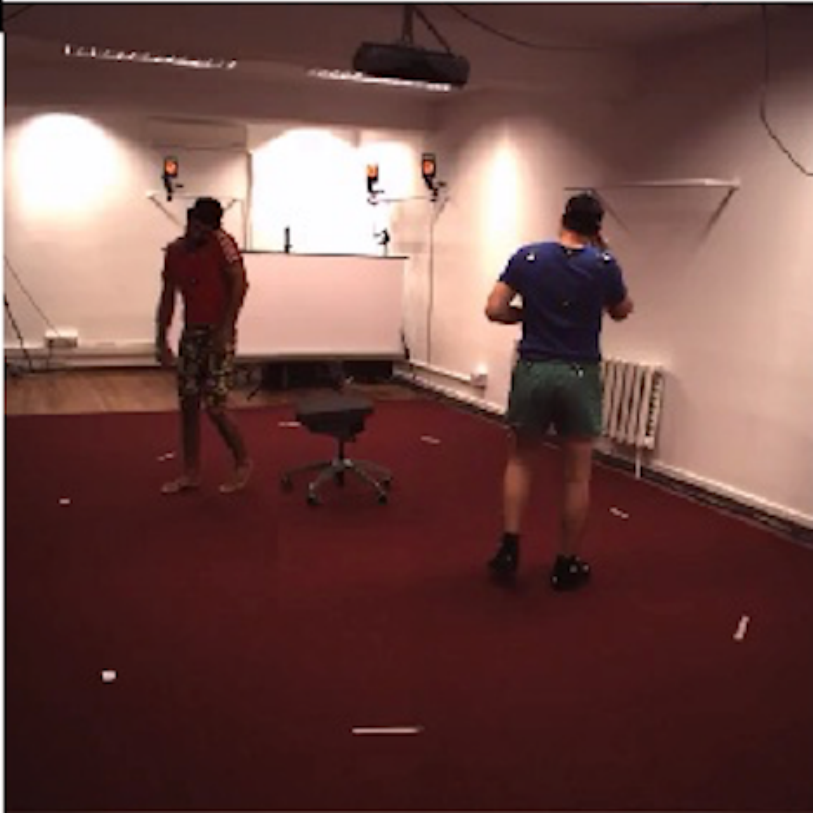} &
\includegraphics[width=0.15\linewidth]{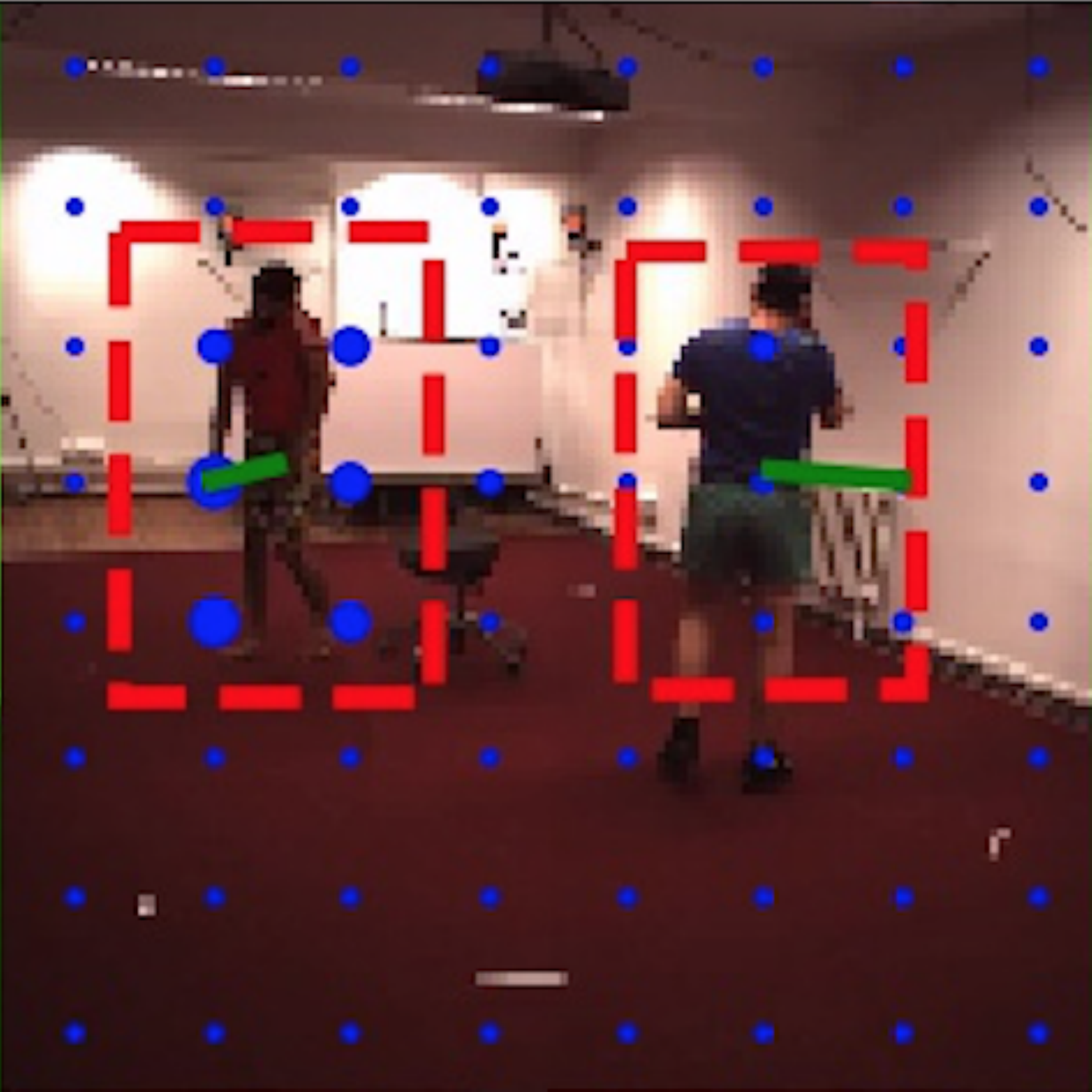} &
\includegraphics[width=0.15\linewidth]{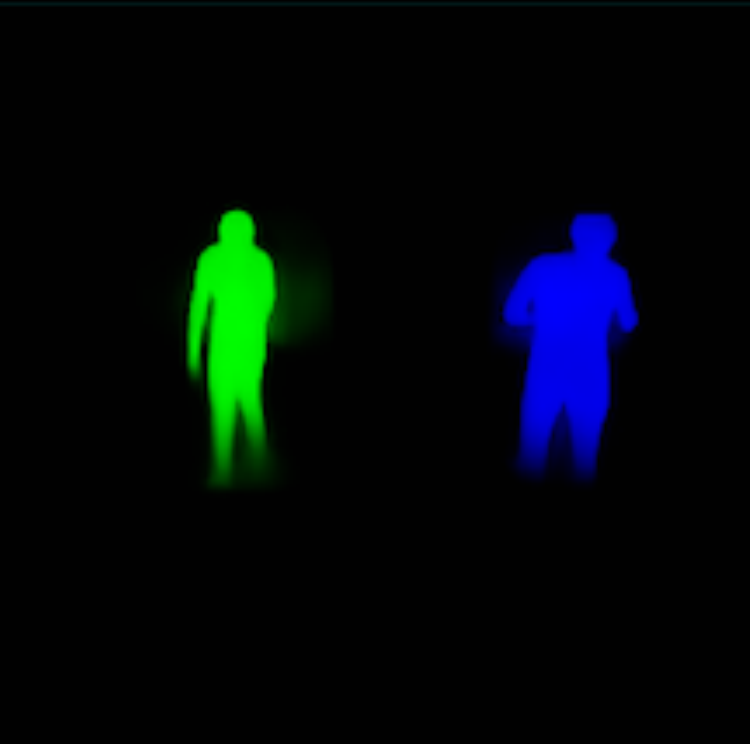} \\ 
{\small (a) Input} & {\small (b) Detection }& {\small (c) Segmentation} & {\small (a) Input} & {\small (b) Detection }& {\small (c) Segmentation}
  \end{tabular}}
  \caption{\textbf{Multi-person detection and segmentation results}, generated by sampling our model multiple times. As the model is trained on single persons this only works for non-intersecting cases.}
  \label{fig:multi_person}
\end{figure}

\parag{Multiple people.} Although our focus is on handling single objects or persons, our probabilistic framework can handle several at test time by sampling more than once. Fig.~\ref{fig:multi_person} shows the predicted cell probability as blue blobs whose size is proportional to it. The fully-convolutional architecture operates locally and thereby predicts a high person probability next to both subjects. As a result, both the detection and segmentation results remain accurate so long persons are sufficient separated.

\subsection{Skiers Filmed Using a PTZ-Cameras}

We now turn to the out-of-the-ordinary motions of six skiers on a slalom course featured in the \ski{} dataset~\cite{Rhodin18a}. The six skiers are split four/one/one to form training, validation, and test sets; totaling to, respectively, 7800, 1818 and 1908 frames. The intrinsic and extrinsic parameters of the pan-tilt-zoom cameras are constantly adjusted to follow the skier. As a result, nothing is static in the images, the background changes quickly, and there are additional people standing in the back. 

\begin{figure}
  \centering
  \begin{tabular}{@{}cccc@{}}

\includegraphics[width=0.25\linewidth]{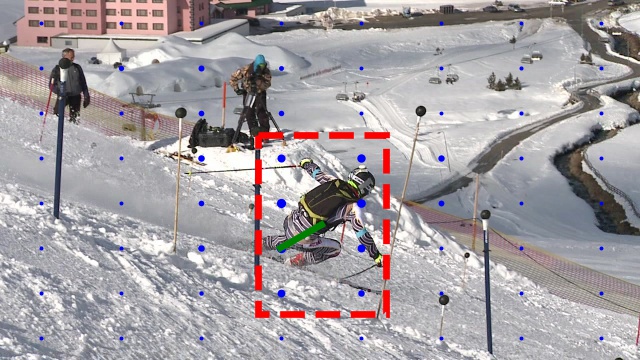} &
\hspace{-0.47cm}
\includegraphics[width=0.25\linewidth]{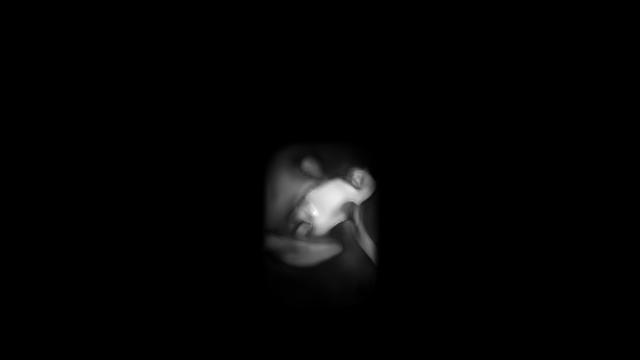}& 
\hspace{-0.47cm}
\includegraphics[width=0.25\linewidth]{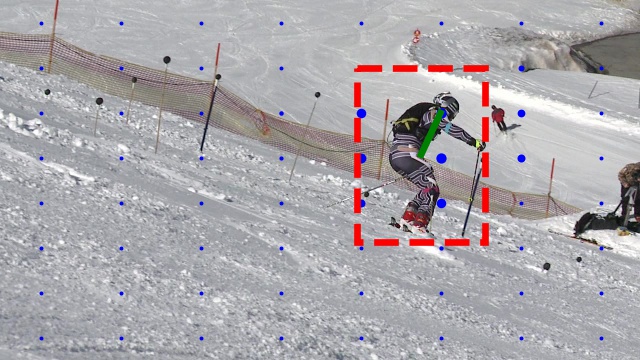} &
\hspace{-0.47cm}
\includegraphics[width=0.25\linewidth]{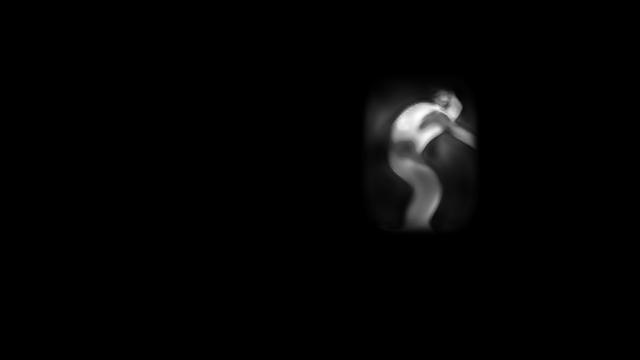} \\

\includegraphics[width=0.25\linewidth]{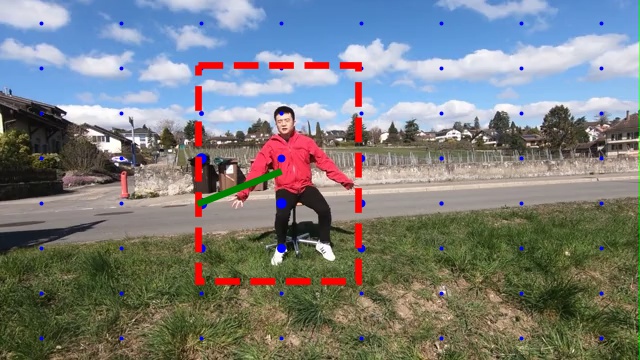} &
\hspace{-0.47cm}
\includegraphics[width=0.25\linewidth]{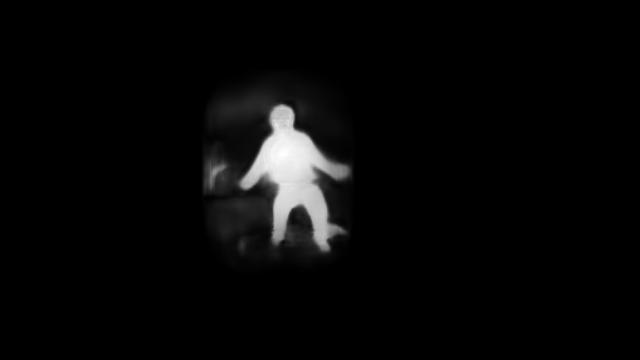}& 
\hspace{-0.47cm}
\includegraphics[width=0.25\linewidth]{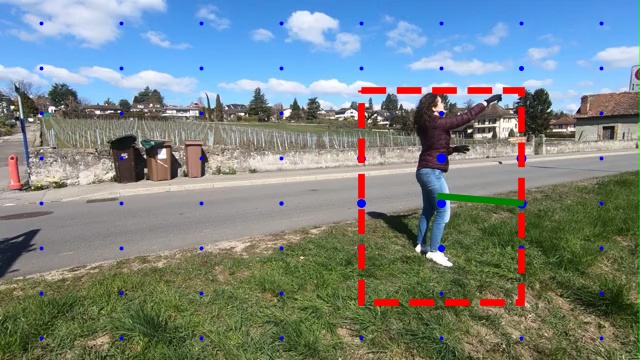} &
\hspace{-0.47cm}
\includegraphics[width=0.25\linewidth]{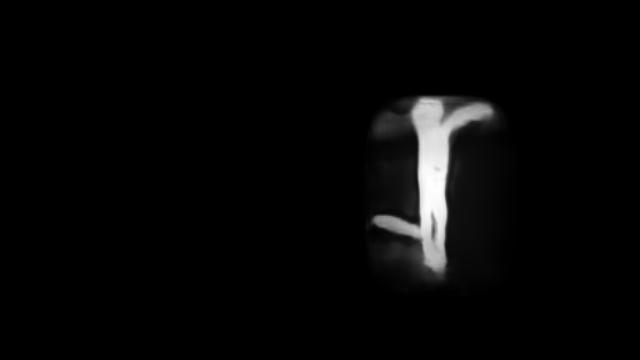} \\

  {\small (a) Detection }&  {\small (b) Segmentation}&  {\small (a) Detection }&  {\small (b) Segmentation}  \vspace{2mm} \\ \\
  \end{tabular}
  \vspace{-5mm}
  \caption{\textbf{Qualitative results on Ski-PTZ-camera and Handheld190k.} Example results on the test images. (a) The detection results show the predicted bounding box with red dashed lines, the relative confidence of the grid cells with blue dots and the bounding box center offset with green lines. (b) Soft segmentation mask predictions. Note that in the second row, the moving clouds are not segmented but the shadow of the person can be included.}
  \label{fig:ski_qual}
\end{figure}

We use the full image as input, evaluate detection accuracy in relation to the available 2D pose annotation, and segmentation accuracy by manually segmenting 36 frames (one from each of the six cameras and two test sequences). As shown on the right side of Table~\ref{tbl:results_hpatches}, our method delivers a mAP$_{0.5}$  score that is significantly better than that of the general YOLO~\cite{Redmon16} detector trained on MSCOCO.  For an analysis of the segmentation quality, we ran several related works on this dataset and list them in Table~\ref{tbl:results_ski_quantitative}, in terms of precision, recall, F-, and J-measure as defined in \cite{Pont17}. To be fair, we compensate for different segmentation masks quantification levels by a grid search (at 0.05 intervals) to select the best threshold in terms of J-measure for each method. This measure is defined as the intersection-over-union between the ground-truth segmentation mask and the prediction. The F-Score is the harmonic average between the precision and the recall on the mask boundaries. 

Interestingly, MaskRCNN trained on a large generic dataset is outperformed by ARP on this dataset. without using any object localization data, our method is on par with MaskRCNN and close to weakly supervised methods that train on large datasets with motion boundary and segmentation mask annotation, while exceeding the self-supervised one of~\cite{Stretcu15}. Our results exceed the only existing self-supervised object segmentation method using deep learning~\cite{Croitoru19} in precision but are slightly behind in recall, F- and J-measures. Part of this difference can be attributed to \cite{Croitoru19} using a segmentation mask discriminator that is trained on the combination of the ImageNet VID and YouTube
Objects datasets. Albeit also trained in a self-supervised fashion, it thereby leverages additional information and results are not one-to-one comparable.

Further qualitative results are shown in Figure~\ref{fig:ski_qual}. The probability distribution, visualized as blue dots that increase in magnitude with the predicted likelihood, show clear peaks on the persons. Limitations are the slightly blurred and bleeding masks and occasional false positives, reducing precision.

\renewcommand{\arraystretch}{1}
\renewcommand{\tabcolsep}{2mm}
\begin{table}
	\centering
\resizebox{0.5\textwidth}{!}{
	\begin{minipage}{0.3\textwidth}
				\centering
		\begin{tabular}{@{}lc@{}}
\multicolumn{2}{c}{\human{} dataset}\\
	\toprule
	Method                         & mAP$_{0.5}$ \\
	\midrule
	NSD \cite{Rhodin19}						& \bf{0.710  }              \\ %
	Ours   					 & {0.580}            \\
	\bottomrule\\
\end{tabular}
\end{minipage}
	\begin{minipage}{0.3\textwidth}
		\centering
\begin{tabular}{@{}lc@{}}
\multicolumn{2}{c}{\ski{}}\\
	\toprule
	Method                         & mAP$_{0.5}$ \\
	\midrule
	YOLOv3~\cite{Redmon16}        					& 0.155                     \\
	Ours   					 & \bf{0.278}            \\
	\bottomrule\\
\end{tabular}
	\end{minipage}} \quad\quad
\vspace{-2mm}	
\caption{\textbf{Detection} results on the \human{} and \ski{} datasets. They are expressed in terms of mAP$_{0.5}$, the mean probability of having an intersection-over-union (IOU) of more than 50\%.} %
\label{tbl:results_h36m}%
\label{tbl:results_hpatches}
\end{table}

\newcommand*\rot{\rotatebox{90}}
\renewcommand{\arraystretch}{1}
\renewcommand{\tabcolsep}{2mm}
\begin{table}
\small
\begin{center}
\resizebox{\columnwidth}{!}{
\begin{tabular}{cc}
\begin{tabular}{@{}lcccc@{}}
&\multicolumn{4}{c}{\ski}\\
\toprule
Method             &Precision &Recall  &F Measure &J Measure\\
\midrule
MaskRCNN \cite{He17} $^d$						  &0.75		&0.65	  &0.68	   &0.65	\\
ARP \cite{Koh17}$^w$										& \bf{0.94}    & \bf 0.76    & \bf{0.83}   & \bf{0.73}      \\
\hline
VideoPCA \cite{Stretcu15}$^s$        					& 0.49     & \bf{0.84}   & 0.61   &  0.56      \\
Unsup-DilateU-Net \cite{Croitoru19}$^s$						&0.74			&0.76		&\bf 0.74 		&\bf0.65	\\
Ours$^s$   									& \bf0.75    & 0.56 &  0.63    &  0.56    \\
\bottomrule
\end{tabular}&

\begin{tabular}{@{}cccc@{}}
\multicolumn{4}{c}{\handheld}\\
\toprule
Precision &Recall  & F measure & J measure\\
\midrule
\bf{0.91}		&\bf 0.82	  &\bf{0.86}	   & \bf{0.77}	\\
0.87    & 0.64    & 0.73   & 0.63      \\
\hline
 0.33     & \bf{0.91}   & 0.48   & 0.49      \\
\bf 0.83	&0.75		& \bf 0.79	& \bf 0.70			\\
 0.75   & 0.74  & 0.74    & 0.66    \\
\bottomrule
\end{tabular}\\
\end{tabular}
}
\end{center}
\caption{\textbf{Segmentation} results on the \ski{} and \handheld{} dataset. Ours exceeds or is on par with the self-supervised methods (marked with $^s$), and approaches the accuracy of weakly-supervised (marked with $^w$) and fully-supervised methods (marked with $^d$).}
\label{tbl:results_ski_quantitative}
\end{table}

\subsection{Daily Activities Captured Using Hand-Held Cameras}

We introduce a new \handheld{} dataset that features three training and two validation sequences that comprise $120\,000$ and $69\,000$ images, respectively, with a single actor performing actions mimicking those introduced in \human.
The camera operators moved laterally, to test robustness to camera translation and hand-held rotation. We provide examples of our detection and segmentation results in Fig.~\ref{fig:ski_qual}, more are given in the supplemental document. Our method is robust to the undirected camera motion and to dynamic background motion,  such as branches swinging in the wind and clouds moving at this windy day, and to salient textures in the background, such as that of the house facade.

To perform a quantitative comparison, we manually segmented 36 validation images taken from six different motion classes----pose, phone, shopping, directions, petting a dog, and greeting----with the subject in many different poses. We then ran several existing methods on this dataset and evaluated the same quantities as for skiing. In this scenario, MaskRCNN yields the highest scores, which is not surprising as the tested sequences are similar to its training set. It is closely followed by \cite{Croitoru19}, which however uses a discriminator that is trained unsupervised on another, larger dataset. Compared to skiing, we exceed ARP~\cite{Koh17} in F- and J-measure and have a larger margin on VideoPCA~\cite{Stretcu15}, both often fail in separating the non-homogeneously moving background due to the hand-held camera motion.

\section{Conclusion}

We have proposed a self-supervised method for object detection and segmentation that lends itself for application in domains where general purpose detectors fail. Our core contributions are the Monte Carlo-based optimization of proposal-based detection, new foreground and background objectives, and their joint training on unlabeled videos captured by static, rotating and handheld cameras. Our latest experiments demonstrate that, even if trained only on single persons, our approach generalizes to multi-person detection, as long as the persons are sufficiently separated.
In contrast to many existing solutions~\cite{Barnich11,Russell14,Croitoru18}, our approach does not exploit temporal cues. In the future, we will integrate temporal dependencies explicitly, which will facilitate addressing the scenario where multiple people interact closely, by incorporating physics-inspired constraints enforcing plausible motion.

\small

\bibliographystyle{ieee}
\bibliography{string,vision,graphics,learning,biomed,misc}

\end{document}